\documentclass[preprint,review,12pt]{elsarticle}

\usepackage{lineno}
\usepackage{graphicx}
\usepackage{amsmath}
\usepackage{amssymb}
\usepackage{setspace}
\usepackage{enumerate}
\usepackage{xcolor}
\usepackage{hyperref}
\usepackage{cleveref}
\definecolor{deepgreen}{rgb}{0,0.5,0}
\hypersetup{
    colorlinks=true,
    citecolor={deepgreen}
}
\usepackage{color}
\usepackage{array}
\usepackage[ruled,linesnumbered,vlined]{algorithm2e}
\usepackage{multirow}
\usepackage{microtype}

\newcolumntype{C}[1]{>{\centering\let\newline\\\arraybackslash\hspace{0pt}}m{#1}}

\journal{Pattern Recognition}

\begin{document}

\begin{frontmatter}

\title{Multi-Classification using One-versus-One Deep Learning Strategy with Joint Probability Estimates}

\author[label1]{Anthony Hei Long Chan\corref{cor1}}
\cortext[cor1]{Corresponding author.}
\author[label2]{Raymond Honfu Chan}
\author[label2]{Lingjia Dai}

\address[label1]{Hong Kong Centre for
Cerebro-Cardiovascular Health Engineering}
\address[label2]{Department of Mathematics, City University of Hong Kong}

\begin{abstract}
The One-versus-One (OvO) strategy is an approach of multi-classification models which focuses on training binary classifiers between each pair of classes. While the OvO strategy takes advantage of balanced training data, the classification accuracy is usually hindered by the voting mechanism to combine all binary classifiers. In this paper, a novel OvO multi-classification model incorporating a joint probability measure is proposed under the deep learning framework. In the proposed model, a two-stage algorithm is developed to estimate the class probability from the pairwise binary classifiers. Given the binary classifiers, the pairwise probability estimate is calibrated by a distance measure on the separating feature hyperplane. From that, the class probability of the subject is estimated by solving a joint probability-based distance minimization problem. Numerical experiments in different applications show that the proposed model achieves generally higher classification accuracy than other state-of-the-art models.
\end{abstract}


\end{frontmatter}

\section{Introduction}

Multi-classification is one of the fundamental tasks in modern image analysis and has a wide variety of real-life applications, including object detection, geography analysis, medical imaging, traffic control system, satellite system, etc. In recent years, deep learning models are mostly implemented to develop effective solutions to most multi-classification problems. However, multi-class problems have a fundamental difference from two-class problems, which causes difficulty to incorporate deep learning models in developing the solutions. To train a high-performance deep neural network, a well-known crucial criterion is to keep the dataset balanced in terms of the number of subjects in different classes. In multi-class problems, such criteria can be hard to achieve.

In most multi-classification models, to solve a $K$-class problem, a classifier is designed or trained to output a $K$-element vector, with each element representing the probability estimate of the subject belonging to the corresponding class respectively. Such an intuitive strategy induces imbalanced data in the training process of the neural networks. In particular, in estimating each element in the output vector, there is only one class to be classified as ``positive'' and all the other classes are classified as ``negative''. The introduction of imbalanced data is inevitable in these models. 

To overcome the difficulty, one may employ One-versus-One (OvO) models instead. For each pair of classes, a binary classifier is developed to classify between them, and it has no discriminating power over all other classes. To label a subject, the results from binary classifiers are combined, conventionally by a simple voting mechanism. OvO models are known to have merit in estimating a more sophisticated decision boundary among classes and hence provide more accurate classification. While OvO models overcome the problem of imbalanced data in the training process, their performance is usually limited by the voting mechanism to combine the binary classifiers. Namely, inaccuracies of each binary classifier are convoluted in the voting mechanism and therefore hinder the final classification accuracy.

In this paper, a novel OvO deep learning model incorporating a joint probability estimate is developed for multi-classification problems. In the proposed model, binary classifiers, which are developed by deep neural networks, are combined by a joint probability algorithm to label the given subject. The joint probability algorithm consists of two steps. Firstly, the binary probability estimate from each classifier is calibrated according to the distance of the subject from the two corresponding classes on the separating feature hyperplane. This step is crucial in adjusting the discriminating power of each classifier, and serves as an extra weighting on the pairwise probability estimate. Afterward, the final class probability is estimated by solving a joint probability problem, that the class probability should satisfy all the calibrated pairwise probabilities as closely as possible. This problem is transformed into a Kullback-Leibler (KL) distance problem, which can be solved by an explicit linear system.

The proposed model has three major merits. Firstly, the posterior probability calibration improves the ultimate classification accuracy by softly inactivating those classifiers which are weak in classifying the given subject between the corresponding two classes. Secondly, the method of combining the pairwise probability estimate by a joint probability measure is novel and generates an accurate and meaningful class probability for the subject. The proposed model is compared with other state-of-the-art multi-classification models on multiple datasets. Numerical experiments under different settings show that the proposed model demonstrates better performance and sets a new benchmark in the multi-classification task. Finally, the proposed model to combine the binary classifiers is mathematically interpretative, hence reliable to maintain the performance at the same level given other datasets, and further modifications to adapt to specific tasks are simple.
 
%\hfill Sep 21, 2022

\section{Previous Work}

Multi-classification models have a long history, and different approaches have been further proposed to suit particular situations. While each of them has its unique advantage, the OvO strategy has been compared with other approaches by researchers \cite{oa_comparison,oa_comparison2,oa_comparison3,oa_comparison4}. Over the years, each approach has been improved by many different techniques.

One major competitor of the OvO strategy is the One-versus-All (OvA) strategy. The OvA strategy is to develop a classifier that estimates the probability of the given subject belonging to each class, and gives the final label to the subject based on the magnitude of each probability \cite{ova_review}. The strategy has been improved by many researchers. For example, a novel hybrid intelligent method is proposed in \cite{ova1} to combine the C4.5 decision tree classifier with the OvA strategy to achieve high accuracy multi-class classification. While the OvA strategy is more intuitive and straightforward to use, it suffers from a major drawback. In training each classifier, there are many more negative samples than positive samples, hence inducing imbalanced training and hindering the classification accuracy. Instead, in OvO models, only two classes are extracted from the database for training a binary classifier each time. Therefore, balanced training is possible by using the OvO strategy, which can achieve a higher classification accuracy.

Compared with the OvA strategy, the OvO strategy breaks down the multi-class problem into many two-class sub-problems. The outputs from each classifier are then combined to label the given subject. The OvO strategy can be incorporated with many two-class classification tools: the SVM, for example. 

Traditionally, Support Vector Machine (SVM) is a powerful tool for two-class tasks \cite{svm,svm2}, and can be easily incorporated with the OvO strategy to deal with multi-class problems \cite{k22}. The SVM-OvO strategy is further improved by different methods. For example, the binary decision tree is incorporated with SVM in \cite{svm_bdt} to achieve multi-class classification. In \cite{svcr}, a new algorithm called Support Vector Classification Regression is proposed using a 1-versus-1-versus-rest strategy. In \cite{svm_error}, the SVM is combined with different classification techniques for comparison and an error-coding algorithm is proposed to replace the usual direction to break down the multi-class problem into two-class problems. In \cite{chan1}, a two-stage method incorporating with the Mumford-Shah model is proposed for multi-class image classification. In \cite{network1}, the SVM is combined with active learning to address unlabeled data and translate an unlabeled multi-classification problem into a classical supervised multi-classification problem. These approaches may be promising in achieving high classification accuracy, especially for a given specific task. However, these approaches usually require a time-consuming fine-tuning of the SVM parameters before deployment.

Other approaches of the OvO strategy have also been investigated. In \cite{ovo1}, an optimizing decision directed acyclic graph (ODDAG) is proposed to improve the accuracy of the OvO strategy by maximizing the fitness on the training set instead of by predefined rules. In \cite{att1,att2}, the OvO strategy is incorporated with attribute learning, in which a class is predicted based on combining the attributes predicted from the network. These strategies may achieve high performance but require more labeling effort.

The modern approach to use the OvO strategy is to incorporate it with neural networks. In \cite{1v1_network1}, an intuitive approach is proposed to train a network for each binary classifier, and then combine the results to achieve multi-class classification. To the best of our knowledge, the most recent and advanced scheme is the one proposed in \cite{1v1_network2}, in which a code matrix is constructed to encode and decode the network output such that it is only required to train one single network to achieve multi-class classification. Apart from consuming less training time, the strategy also achieves higher accuracy than the one proposed in \cite{1v1_network1} when compared over many datasets. 

It is noteworthy that existing OvA models can be modified into OvO models to further improve the accuracy. In \cite{ecg}, an OvA deep neural network is proposed and trained on millions of subjects for disease classification. In the later part of this paper,  we will see how we incorporate the OvA models with the OvO strategy to improve the model's accuracy.

\section{Method}
In this section, our proposed One-versus-One strategy is explained in detail. It consists of three steps:
\begin{enumerate}
    \item Acquiring binary classifiers by training neural networks,
    \item Posterior probability calibration to apply weighting on the classifiers,
    \item Joint probability estimate to label the subject.
\end{enumerate}

\subsection{Binary classifiers}

Concerning the $K$-class classification problem, the conventional multi-classification strategy focuses on designing a single network ${\cal D}$ to provide an end-to-end solution. That is, given a subject $x$ from the database $S$, ${\cal D}(x)=[{\cal D}_1(x),\dots,{\cal D}_K(x)]$ is a $K$-vector with each node approximating the probability of $x$ belonging to the corresponding class. 
Assuming that each class has $n$ data,
in training the network, each node ${\cal D}_i$ from ${\cal D}$ is presented to $n$ positive data and $(K-1)\times n$ negative data, inducing imbalanced data training and may hinder the classification accuracy. 

In this paper, we propose to use binary classifiers to overcome the imbalanced data issue. There are two methods to acquire binary classifiers. They are as follows.

In the One-versus-One (OvO) strategy, the $K$-class problem is decomposed into 
% $C^K_2=K(K-1)/2$ 
$\binom K 2=K(K-1)/2$ 
binary classification problems to classify between each pair of the classes $i$ and $j$, $1\leq i< j\leq K$. Each binary problem is solved by a binary classification network ${\cal D}^j_i$, called a binary classifier, which reports a binary class probability estimate ${\cal D}^j_i(x)$ given a subject $x$. Here, ${\cal D}^j_i(x)$ is close to $1$ if $x$ belongs to class $i$, and is close to $0$ if $x$ belongs to class $j$. Each classifier can be trained independently from scratch on the corresponding subset of the given database. This promotes balance training and improves the overall classification accuracy. 

Alternatively, if we are already given an OvA network ${\cal D}$, it can be modified into multiple binary classifiers as follows. For each class $j$, the network ${\cal D}^j$ is trained from ${\cal D}$ on the subset $S_j$ of the database $S$ which contains only the subjects belonging to the class $j$. Therefore, each node ${\cal D}^j_i$ from ${\cal D}^j(x)=[{\cal D}^j_1(x),\dots,{\cal D}^j_K(x)]$ is a binary classifier such that ${\cal D}^j_i(x)$ is close to $1$ if $x$ belongs to class $i$, and is close to $0$ if $x$ belongs to class $j$. In this manner, all ${\cal D}^j_i$'s compose a complete set of binary classifiers for the $K$ classes for $1\leq i< j\leq K$. 

Both methods to build binary classifiers from scratch or from existing models can be applied to our proposed OvO strategy and are tested by different experiments, as demonstrated in the next section.

\subsection{Posterior probability calibration}

The conventional OvO strategy employs a voting mechanism to combine all the binary classification results. 
More formally, the class label of the subject $x$ is determined by
\begin{equation}
    C(x)=\text{argmax}_i[\sum_{j:j\neq i}I_{{\cal D}^j_i(x) > {\cal D}^i_j(x)}],
\label{argmax}
\end{equation}
% \begin{equation}
%     C=\text{argmax}_i[\sum_{j:j\neq i}I_{\gamma_{ij}(x)>\gamma_{ji}(x)}],
% \label{argmax}
% \end{equation}
% \begin{equation}
%     C=\text{argmax}_i[\sum_{j:j\neq i}I_{\gamma_{ij}>\gamma_{ji}}],
% \label{argmax}
% \end{equation}
where $I_S$ is the indicator function such that $I_S=1$ if the event $S$ is true, and $I_S=0$ otherwise. The pairwise class probability estimate ${\cal D}^j_i(x)$ is reported by the binary classifier ${\cal D}^j_i$ of the class $i$ and $j$.
Let $p_i$ denotes the unknown class probability of $x$ belonging to the class $i$ among all $K$ classes, ${\cal D}^j_i(x)$ is the estimate of $\mu_{ij}=p_i/(p_i+p_j)$.
% is unknown for binary classifiers in the OvO strategy.
% The class probability $\mathbf{p}=(p_1,p_2,\dots,p_K)$ for a $K$-class problem can not be accessed by binary classifiers directly in the OvO strategy.
% and $\gamma_{ij} = p_i/(p_i+p_j)$,
Such a voting mechanism is accurate only if all the binary classifiers have fair sensitivity to the given subject. Inspired by \cite{k22,prob_merge1}, in our proposed model, the decision module is developed based on joint probability calibration to improve the accuracy. 
% We explore the estimate of the class probability $\mathbf{p}=(p_1,p_2,\dots,p_K)$ before getting the label of $x$.

% \textcolor{blue}{When the OvO strategy is incorporated with neural networks, the binary classifier ${\cal D}^j_i$ is acquired by training network on 
% the pairwise class probability $\gamma_{ij}(x)$ of the subject $x$ is also denoted as ${\cal D}^j_i(x)$ from the binary classification network ${\cal D}^j_j$.} 
The pairwise probability estimate ${\cal D}^j_i(x)$ is calibrated by applying an extra weighting based on the sensitivity of the binary classifier ${\cal D}^j_i$ to classify $x$ between the two corresponding classes. Mathematically, the calibration is designed to be proportional to the distance of the subject from the two classes on the separating feature hyperplane, which is given by the sigmoid function,
\begin{equation}
r_{ij}(x)=P_{\eta,\tau}({\cal D}^j_i(x))=\frac{1}{1+e^{\eta {\cal D}^j_i(x)+\tau}}.
\label{rij}
\end{equation}
Here, $\eta$ and $\tau$ are two parameters to control the distance measurement on the separating feature hyperplane.

According to \cite{prob_merge1}, the best parameters $\eta$ and $\tau$ are determined by solving the following negative log-likelihood problem:
\begin{equation}
\begin{aligned}
\min F(\eta,\tau)&=-\sum_{k}\big(t_k\log(r_{ij}(x_k))+(1-t_k)\log(1-r_{ij}(x_k))\big), \\
\text{where } 
t_k&=\left\{
        \begin{aligned}
       \frac{N_i+1}{N_i+2} & \text{  if } y_k=i, \\
        \frac{1}{N_j+2} & \text{  if } y_k=j,
        \end{aligned}
        \right.
\end{aligned}
\end{equation}
% \begin{equation}
% \begin{aligned}
% \min F(\eta,\tau)&=-\sum_{k}(t_k\log(P_{\eta,\tau}({\cal D}^j_i(x_k))))+(1-t_k)\log(1-P_{\eta,\tau}({\cal D}^j_i(x_k))))), \\
% \text{where } 
% t_k&=\left\{
%         \begin{aligned}
%        \frac{N_i+1}{N_i+2} & \text{  if } y_k=i, \\
%         \frac{1}{N_j+2} & \text{  if } y_k=j,
%         \end{aligned}
%         \right.
% \end{aligned}
% \end{equation}
over all the training samples $x_k$ labeled by $y_k$, $k=1,\cdots,N_i+N_j$, where $N_i$ and $N_j$ are the number of samples in the class $i$ and $j$ respectively. 
Finally, all calibrated binary probability estimates $r_{ij}(x)$ will be combined to give the final labeling to the given subject $x$.

\subsection{Class labeling by joint probability estimate}

In \cite{pairwise_coupling}, it is proved that pairwise class probabilities can be combined into a joint probability measure to increase labeling accuracy. Intuitively,  given a subject $x$, our task is to find a class probability $\mathbf{p}=(p_1,p_2,\dots,p_K)$ such that $\mu_{ij}=p_i/(p_i+p_j)$ is close to the binary probability estimate $r_{ij}=r_{ij}(x)$ as defined in \Cref{rij}. Such a $\mathbf{p}$ can be found by solving the following Kullback-Leibler (KL) distance minimization problem:
% \begin{equation}
%     l(\mathbf{p})=\sum_{i\neq j}n_{ij}r_{ij}\text{log}(r_{ij}/\gamma_{ij}),
% \end{equation}
% subject to 
% \begin{equation}
%     \sum^K_{i=1}p_i=1,\quad p_i>0,\quad i=1,2,\dots,K,
% \end{equation}
\begin{equation}
\begin{aligned}
     &l(\mathbf{p})=\sum_{i\neq j}n_{ij}r_{ij}\text{log}(r_{ij}/\mu_{ij}),\\
    &\text{s.t. }     \sum^K_{i=1}p_i=1,\quad p_i\geq 0,\quad i=1,2,\dots,K,
\end{aligned}
\end{equation}
where $n_{ij}=N_i+N_j$ is the number of subjects in the training set. For more details about the KL distance minimization problem, readers are referred to \cite{pairwise_coupling} for more insight.

Inspired by \cite{k22}, instead of solving the above KL distance minimization problem directly, we transform it as follows to achieve better performance:
% it can be transformed as follows to achieve better performance:

\begin{equation} 
\label{eq:prob_vec}
\begin{aligned}
    &\min_{\mathbf{p}} \frac{1}{2} \sum_{1\leq i<j\leq K}(r_{ij}p_j-r_{ji}p_i)^2,\\
    &\text{s.t. } \sum_{i=1}^{K}p_{i}=1, \quad p_i\geq 0, \quad i=1,2,\dots,K .
\end{aligned}
\end{equation}
The optimal solution to the above problem can be determined by solving the linear system
\begin{equation}
    \begin{bmatrix} Q & \mathbf{e} \\ \mathbf{e}^\intercal & 0 \end{bmatrix}
    \begin{bmatrix} \mathbf{p} \\ b \end{bmatrix}
    =
    \begin{bmatrix} \mathbf{0} \\ 1 \end{bmatrix},
\end{equation}
where
\begin{equation}\nonumber
Q_{i,j}=\left\{
        \begin{aligned}
        \sum_{s\neq i} r_{si}^2 & \text{  if } i=j, \\
        -r_{ji}r_{ij} & \text{  if } i\neq j,
        \end{aligned}
        \right.
\end{equation}
and $b$ is the Lagrange multiplier of the equality constraint in \eqref{eq:prob_vec}, $\mathbf{e}$ is the all-ones vector, and $\mathbf{0}$ is the all-zeros vector. 

The probability vector $\mathbf{p}$ measures the probability of the given subject $x$ belonging to each class. Each class probability measure is compatible with the classification results of all the binary classifiers, and is calibrated such that a suitable weighting is applied according to the classifier's sensitivity to the given subject. Finally, the sample $x$ can be labeled as class $C$ where
\begin{equation}
    C={\arg\max}_{i} \mathbf{p}=\arg\max_{i} [p_1, p_2,..., p_K],
\end{equation}
when each subject belongs to only one class. Otherwise, a simple thresholding can be applied to each node of $\mathbf{p}$ to give multiple labels to $x$.

\section{Experiments}

The proposed OvO strategy can be applied to many different tasks. In this paper, experiments are demonstrated on two different applications, namely, in computer vision and in medical applications. 

\subsection{Multi-class Image Classification}

In computer vision, multi-class image classification is a well-studied yet challenging task. Conventional multi-classification strategy suffers from the disadvantage of imbalanced training if there are many classes to separate. In below, an experiment is done to compare the performance among traditional multi-classification methods, other OvO methods, and our proposed OvO method.

The proposed model is compared with those recorded in \cite{1v1_network2}, in which the author compared two classification methods, the one-versus-one (OvO) strategy they proposed and the conventional one-versus-all (OvA) strategy on different datasets, with different neural network settings. 

\begin{figure}[h]
    \centering
    \includegraphics[width=\textwidth]{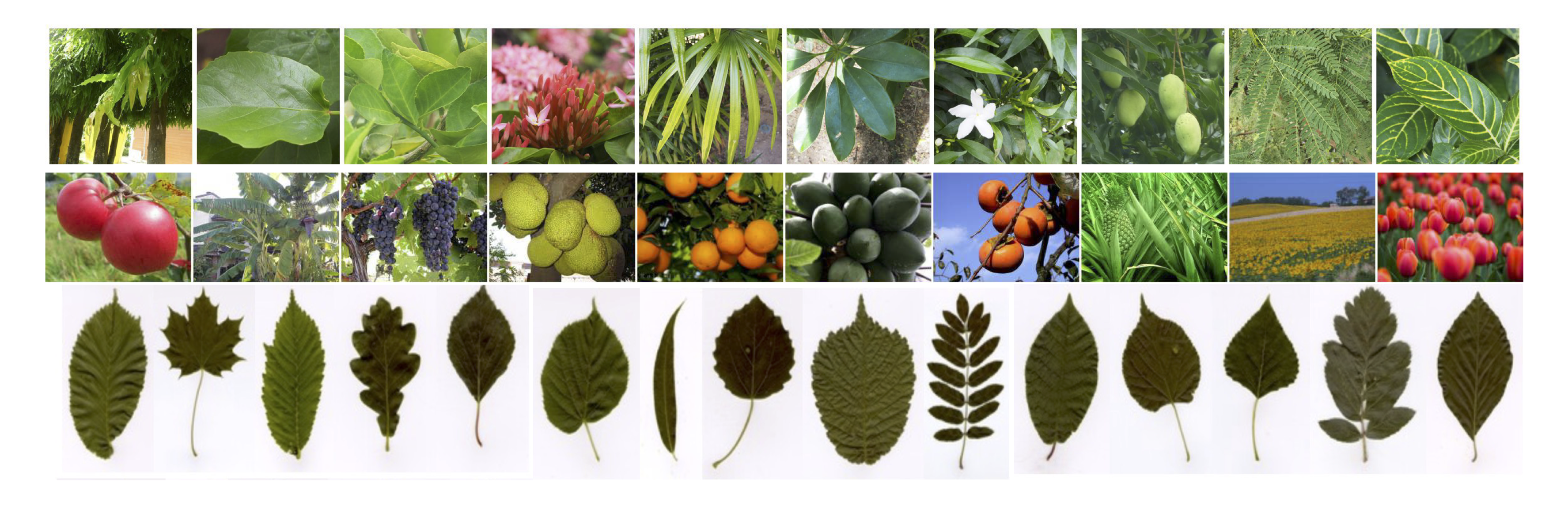}
    \caption{Some example images from the datasets. First row: AgrilPlant images; Middle row: Tropic images; Bottom row: Swedish images}
    \label{fig:dataset_image}
\end{figure}

For fairness, the performance of our proposed model is compared on the same datasets as in \cite{1v1_network2}. In particular, three datasets are used, namely, the AgrilPlant dataset \cite{agrilplant}, the Tropic dataset \cite{1v1_network2}, and the Swedish dataset \cite{swedish} respectively. The composition of each dataset is recorded in \Cref{tab:dataset_image} and some example images are shown in \Cref{fig:dataset_image}. For more details about the datasets, readers are referred to \cite{1v1_network2}.

\begin{table}[h]
    \centering
    \caption{Composition of each dataset used in the multi-class image classification experiment}
    \label{tab:dataset_image}
    \begin{tabular}{c|ccc}
        Dataset & No. of images & No. of classes & No. of images per class \\
        \hline
        AgrilPlant & 3000 & 10 & 300 \\
        Tropic & 5276 & 20 & 221--371  \\
        Swedish & 1125 & 15 & 75  \\
    \end{tabular}

\end{table}

In order to analyze the models in a more sophisticated way, the experiments are performed under the following settings, including:
\begin{enumerate}
    \item the choice of neural networks (ResNet-50 and Inception-V3),
    \item the training schemes of neural networks (from scratch, or fine-tune from pre-trained networks),
    \item the number of classes involved (5 classes or 10 classes, randomly chosen from the corresponding datasets), and
    \item the train size from 10\% -- 100\% of the entire dataset.
\end{enumerate}
For the last point, \Cref{tab:trainsize} records the number of training images per class with different training sizes for the datasets. 

\begin{table}[h]
    \centering
    \caption{Number of training images per class involved in different sub-sampling settings of the datasets}
    \label{tab:trainsize}
    \begin{tabular}{c|ccc}
        Train size & \multicolumn{3}{c}{Dataset} \\
        (\%) & AgrilPlant & Tropic & Swedish  \\
        \hline
        10 & 24 & 15--26 & 2--3 \\
        20 & 48 & 31--52 & 5  \\
        50 & 120 & 77--130 & 12--13  \\
        80 & 192 & 124--207 & 20  \\
        100 & 240 & 155--259 & 25 \\
    \end{tabular}
\end{table}

The experimental results are investigated here. 
% Firstly, we consider the comparison using neural networks refined from pre-trained networks. 
Firstly, we consider the comparison using the two neural networks, ResNet-50 and Inception-V3, refined from pre-trained models. The neural networks are initialized with the pre-trained weights from the ImageNet dataset.
The classification results, including average accuracy and standard deviation, are recorded in \Cref{tab:resnet_refine,tab:inception_refine}, respectively. 
To enhance the readability, we have highlighted the numbers in bold that represent the highest accuracy in every row of the tables.
% achieved by each method 

\begin{table}[!t]
    \footnotesize
    \centering
    \caption{Classification performance (average accuracy and standard deviation) of the fine-tuned ResNet-50 combined with the three different classification methods. The bold numbers indicate the highest accuracy in each case }
    \label{tab:resnet_refine}
    \resizebox{0.70\textwidth}{!}{
    \begin{tabular}{c|ccc}
        \multicolumn{4}{c}{Refined ResNet-50}\\
        \hline
        Train Size & \multicolumn{3}{c}{AgrilPlant5} \\
        (\%) & Proposed & OvO\cite{1v1_network2} & OvA\cite{1v1_network2}\\
        \hline
        10 & \textbf{93.40} $\pm$ 1.58 & 91.13 $\pm$ 1.39 & 89.47 $\pm$ 3.03\\
        20 & \textbf{96.68} $\pm$ 1.08 & 93.93 $\pm$ 2.47 & 92.40 $\pm$ 1.16\\
        50 & \textbf{97.20} $\pm$ 1.12 & 96.33 $\pm$ 1.62 & 96.07 $\pm$ 0.64\\
        80 & \textbf{97.86} $\pm$ 1.66 & 97.27 $\pm$ 0.86 & 97.07 $\pm$ 1.34\\
        100 & \textbf{99.42} $\pm$ 0.44 & 97.60 $\pm$ 1.44 & 97.33 $\pm$ 1.33\\
        \hline
        \hline
        Train Size & \multicolumn{3}{c}{AgrilPlant10} \\
        (\%) & Proposed & OvO\cite{1v1_network2} & OvA\cite{1v1_network2}\\
        \hline
        10 & \textbf{94.27} $\pm$ 0.66 & 93.13 $\pm$ 1.57 & 93.17 $\pm$ 0.31\\
        20 & 95.77 $\pm$ 0.73 & 95.83 $\pm$ 1.87 & \textbf{96.17} $\pm$ 0.87\\
        50 & \textbf{97.80} $\pm$ 0.38 & 97.73 $\pm$ 1.11 & 97.67 $\pm$ 0.94\\
        80 & \textbf{98.53} $\pm$ 0.14 & 98.40 $\pm$ 0.48 & 98.47 $\pm$ 0.40\\
        100 & \textbf{98.67} $\pm$ 0.41 & 98.47 $\pm$0.70 & 98.63 $\pm$ 0.70\\
        \hline
        \hline
        Train Size & \multicolumn{3}{c}{Tropic5} \\
        (\%) & Proposed & OvO\cite{1v1_network2} & OvA\cite{1v1_network2}\\
        \hline
        10 & \textbf{96.83} $\pm$ 0.93 & 96.80 $\pm$ 1.45 & 96.61 $\pm$ 1.20\\
        20 & \textbf{98.21} $\pm$ 1.16 & 98.16 $\pm$ 0.88 & 97.87 $\pm$ 1.09\\
        50 & 99.37 $\pm$ 0.13 & \textbf{99.52} $\pm$ 0.38 & 99.22 $\pm$ 0.47\\
        80 & 99.52 $\pm$ 0.33 & \textbf{99.66} $\pm$ 0.37 & 99.56 $\pm$ 0.32\\
        100 & 99.71 $\pm$ 0.32 & 99.66 $\pm$ 0.28 & \textbf{99.76} $\pm$ 0.24\\
        \hline
        \hline
        Train Size & \multicolumn{3}{c}{Tropic10} \\
        (\%) & Proposed & OvO\cite{1v1_network2} & OvA\cite{1v1_network2}\\
        \hline
        10 & \textbf{94.29} $\pm$ 0.81 & 92.54 $\pm$ 1.91 & 91.96 $\pm$ 1.20\\
        20 & 96.71 $\pm$ 0.64 & 95.80 $\pm$ 0.89 & \textbf{97.70} $\pm$ 0.30\\
        50 & 99.17 $\pm$ 0.34 & 98.72 $\pm$ 0.29 & \textbf{99.19} $\pm$ 0.17\\
        80 & \textbf{99.48} $\pm$ 0.19 & 99.24 $\pm$ 0.28 & 99.41 $\pm$ 0.25\\
        100 & \textbf{99.82} $\pm$ 0.25 & 99.58 $\pm$ 0.11 & 99.71 $\pm$ 0.17\\
        \hline
        \hline
        Train Size & \multicolumn{3}{c}{Swedish5} \\
        (\%) & Proposed & OvO\cite{1v1_network2} & OvA\cite{1v1_network2}\\
        \hline
        10 & \textbf{94.73} $\pm$ 6.08 & 90.48 $\pm$ 4.79 & 89.68 $\pm$ 6.14\\
        20 & \textbf{98.00} $\pm$ 2.44 & 97.44 $\pm$ 1.85 & 98.08 $\pm$ 2.14\\
        50 & 98.55 $\pm$ 2.77 & \textbf{99.76} $\pm$ 0.36 & 99.60 $\pm$ 0.28\\
        80 & \textbf{99.82} $\pm$ 0.41 & 99.76 $\pm$ 0.36 & 99.82 $\pm$ 0.18\\
        100 & \textbf{100.00} $\pm$ 0.00 & 99.92 $\pm$ 0.18 & 99.92 $\pm$ 0.18\\
        \hline
        \hline
        Train Size & \multicolumn{3}{c}{Swedish10} \\
        (\%) & Proposed & OvO\cite{1v1_network2} & OvA\cite{1v1_network2}\\
        \hline
        10 & \textbf{97.82} $\pm$ 1.33 & 90.40 $\pm$ 2.37 & 87.88 $\pm$ 1.88\\
        20 & 98.64 $\pm$ 0.45 & \textbf{98.76} $\pm$ 0.96 & 96.80 $\pm$ 2.04\\
        50 & 99.55 $\pm$ 0.32 & 99.60 $\pm$ 0.20 & \textbf{99.72} $\pm$ 0.23\\
        80 & 99.64 $\pm$ 0.59 & \textbf{99.92} $\pm$ 0.18 & 99.68 $\pm$ 0.39\\
        100 & 99.82 $\pm$ 0.25 & \textbf{99.92} $\pm$ 0.11 & 99.92 $\pm$ 0.18
    \end{tabular}
    }
\end{table}

\begin{table}[!t]
    \footnotesize
    \centering
    \caption{Classification performance (average accuracy and standard deviation) of the fine-tuned Inception-V3 combined with the three different classification methods. The bold numbers indicate the highest accuracy in each case }
    \label{tab:inception_refine}
    \resizebox{0.70\textwidth}{!}{
    \begin{tabular}{c|ccc}
        \multicolumn{4}{c}{Refined Inception-V3}\\
        \hline
        Train Size & \multicolumn{3}{c}{AgrilPlant5} \\
        (\%) & Proposed & OvO\cite{1v1_network2} & OvA\cite{1v1_network2}\\
        \hline
        10 & \textbf{94.80} $\pm$ 2.39 & 88.67 $\pm$ 2.13 & 90.40 $\pm$ 2.42\\
        20 & \textbf{98.20} $\pm$ 0.90 & 92.27 $\pm$ 2.09 & 92.07 $\pm$ 1.86\\
        50 & \textbf{99.00} $\pm$ 0.62 & 96.20 $\pm$ 1.66 & 96.27 $\pm$ 1.44\\
        80 & \textbf{99.87} $\pm$ 0.18 & 96.27 $\pm$ 1.16 & 97.53 $\pm$ 0.69\\
        100 & \textbf{100.00} $\pm$ 0.00 & 97.00 $\pm$ 1.18 & 97.07 $\pm$ 1.23\\
        \hline
        \hline
        Train Size & \multicolumn{3}{c}{AgrilPlant10} \\
        (\%) & Proposed & OvO\cite{1v1_network2} & OvA\cite{1v1_network2}\\
        \hline
        10 & 93.80 $\pm$ 0.62 & 92.13 $\pm$ 1.52 & \textbf{94.87} $\pm$ 0.88\\
        20 & 96.20 $\pm$ 0.57 & 94.47 $\pm$ 1.77 & \textbf{96.67} $\pm$ 0.59\\
        50 & \textbf{98.03} $\pm$ 0.34 & 97.13 $\pm$ 1.02 & 98.03 $\pm$ 0.77\\
        80 & 98.33 $\pm$ 0.42 & 97.93 $\pm$ 0.51 & \textbf{98.77} $\pm$ 0.57\\
        100 & \textbf{98.87} $\pm$ 0.14 & 98.07 $\pm$ 0.56 & 98.83 $\pm$ 0.53\\
        \hline
        \hline
        Train Size & \multicolumn{3}{c}{Tropic5} \\
        (\%) & Proposed & OvO\cite{1v1_network2} & OvA\cite{1v1_network2}\\
        \hline
        10 & 96.47 $\pm$ 1.57 & \textbf{97.15} $\pm$ 1.72 & 96.61 $\pm$ 2.50\\
        20 & \textbf{98.81} $\pm$ 1.15 & 97.39 $\pm$ 1.22 & 98.74 $\pm$ 0.99\\
        50 & \textbf{99.66} $\pm$ 0.22 & 99.32 $\pm$ 0.32 & 99.47 $\pm$ 0.56\\
        80 & \textbf{99.90} $\pm$ 0.13 & 99.66 $\pm$ 0.13 & 99.61 $\pm$ 0.22\\
        100 & \textbf{99.90} $\pm$ 0.13 & 99.76 $\pm$ 0.24 & 99.81 $\pm$ 0.32\\
        \hline
        \hline
        Train Size & \multicolumn{3}{c}{Tropic10} \\
        (\%) & Proposed & OvO\cite{1v1_network2} & OvA\cite{1v1_network2}\\
        \hline
        10 & 93.92 $\pm$ 0.86 & 92.93 $\pm$ 1.21 & \textbf{94.60} $\pm$ 1.52\\
        20 & 97.39 $\pm$ 1.17 & 96.01 $\pm$ 0.98 & \textbf{98.25} $\pm$ 0.57\\
        50 & 99.35 $\pm$ 0.24 & 98.75 $\pm$ 0.27 & \textbf{99.53} $\pm$ 0.41\\
        80 & 99.66 $\pm$ 0.24 & 99.32 $\pm$ 0.23 & \textbf{99.79} $\pm$ 0.15\\
        100 & 99.74 $\pm$ 0.21 & 99.56 $\pm$ 0.22 & \textbf{99.87} $\pm$ 0.16\\
        \hline
        \hline
        Train Size & \multicolumn{3}{c}{Swedish5} \\
        (\%) & Proposed & OvO\cite{1v1_network2} & OvA\cite{1v1_network2}\\
        \hline
        10 & \textbf{98.91} $\pm$ 1.00 & 94.88 $\pm$ 4.10 & 92.48 $\pm$ 4.23\\
        20 & \textbf{99.64} $\pm$ 0.50 & 97.44 $\pm$ 3.26 & 97.52 $\pm$ 3.06\\
        50 & \textbf{99.82} $\pm$ 0.41 & 99.68 $\pm$ 0.18 & 99.68 $\pm$ 0.04\\
        80 & \textbf{99.98} $\pm$ 0.04 & 99.92 $\pm$ 0.18 & 99.92 $\pm$ 0.18\\
        100 & \textbf{100.00} $\pm$ 0.00 & 99.92 $\pm$ 0.18 & 99.92 $\pm$ 0.18\\
        \hline
        \hline
        Train Size & \multicolumn{3}{c}{Swedish10} \\
        (\%) & Proposed & OvO\cite{1v1_network2} & OvA\cite{1v1_network2}\\
        \hline
        10 & \textbf{97.91} $\pm$ 1.66 & 84.56 $\pm$ 2.56 & 91.72 $\pm$ 4.44\\
        20 & \textbf{99.09} $\pm$ 0.32 & 97.68 $\pm$ 1.40 & 98.96 $\pm$ 0.71\\
        50 & \textbf{99.91} $\pm$ 0.20 & 99.72 $\pm$ 0.11 & 99.84 $\pm$ 0.17\\
        80 & \textbf{99.91} $\pm$ 0.20 & 99.76 $\pm$ 0.17 & 99.88 $\pm$ 0.11\\
        100 & \textbf{100.00} $\pm$ 0.00 & 99.92 $\pm$ 0.11 & 99.92 $\pm$ 0.18
    \end{tabular}
    }
\end{table}

Incorporated with the refined ResNet-50, it is observed that our proposed model has a generally better performance than the others. 

\begin{itemize}
    \item On the AgrilPlant dataset, the proposed OvO model outperforms the other models. If 5 classes are randomly chosen, when the training size is small (i.e. $10\%$), our model has $2.27\%$ and $3.93\%$ advantage in mean accuracy over the OvO model and the OvA model presented in \cite{1v1_network2} respectively. The advantage goes to $1.86\%$ and $2.09\%$ when the training size goes up to $100\%$.  If 10 classes are chosen instead, the advantage of our model is $1.14\%$ and $1.1\%$ respectively when the training size is as small as $10\%$, and $0.2\%$ and $0.04\%$ respectively when all the training data is used.
    \item On the Tropic dataset, the proposed model has fair performance compared with other models. If 5 classes are chosen, when the training size is $10\%$, our model has $0.03\%$ and $0.22\%$ advantage in mean accuracy respectively. The advantage goes to $0.05\%$ when the training size goes up to $100\%$, compared with the OvO model presented in \cite{1v1_network2}. However, in this particular setting, even the original OvO model is outperformed by the OvA model. If the 10 classes are picked, the advantage of our model is $1.75\%$ and $2.33\%$ respectively when the training size is $10\%$, and $0.24\%$ and $0.11\%$ respectively when the training size is $100\%$.
    \item On the Swedish dataset, the proposed model outperforms the others when 5 classes are chosen. It achieves $100\%$ accuracy if the training size is $100\%$ and has a $4.25\%$ and $5.05\%$ advantage in mean accuracy when the training size is $10\%$. If 10 classes are considered instead, the proposed model outperforms other models by $7.42\%$ and $9.94\%$ mean accuracy when $10\%$ training size is used. However, if more training data is included, the proposed model has a slight disadvantage over the OvO model in \cite{1v1_network2}.
\end{itemize}

To conclude the results of the experiments under the refined ResNet50 setting, it is observed that our proposed model has a general advantage over other models, with exceptions under a few specific settings.

The following paragraph evaluates the experiment results when the models are incorporated with the refined Inception-V3 network.

\begin{itemize}
    \item On the AgrilPlant dataset, the proposed model outperforms the other models in most settings. If 5 classes are chosen, our proposed model outperforms other models in mean accuracy by $6.13\%$ and $4.4\%$ when compared with the OvO and OvA models respectively, when the training size is $10\%$. When the training size is $100\%$, our proposed model achieves $100\%$ accuracy and outperforms other models by $3\%$ and $2.93\%$ respectively. If 10 classes are chosen, when the training size is $10\%$, the best model is the OvA model instead, in which our proposed model has $1.07\%$ disadvantage. However, our proposed model outperforms other models by $0.8\%$ and $0.04\%$ respectively, when the training size is $100\%$.
    \item On the Tropic dataset, the proposed model performs the best if 5 classes are chosen. As the training size increases, our model outperforms the others by as much as $0.14\%$ and $0.09\%$ respectively, when the training size is $100\%$. If 10 classes are picked, the OvA model has a stably better performance.
    \item On the Swedish dataset, the proposed model outperforms the other models in all settings. It achieves $100\%$ accuracy no matter if 5 or 10 classes are chosen, when the training size is $100\%$. When the training size is $10\%$, our model outperforms the others by $4.03\%$ and $6.43\%$ respectively if 5 classes are picked, and by $13.35\%$ and $6.19\%$ respectively if 10 classes are picked.
\end{itemize}

From \Cref{tab:resnet_refine,tab:inception_refine}, it is observed that our proposed model performs slightly worse on the Tropic dataset, but with a great advantage on the AgrilPlant dataset and on the Swedish dataset. It is noteworthy that in classifying the Tropic dataset, most of the time the OvA model performs better; and the Tropic dataset is the only dataset with uneven distribution of the number of samples in each class. Therefore, it is possible that when the dataset is not evenly distributed, the discriminating power of the OvO strategy may be hindered.

The evaluation process proceeds by considering models that incorporate networks trained from scratch. In contrast to using pre-trained weights, these models have their network weights initialized randomly. The results are recorded in \Cref{tab:resnet_scratch,tab:inception_scratch} respectively. 

\begin{table}[!t]
    \footnotesize
    \centering
    \caption{Classification performance (average accuracy and standard deviation) of the ResNet-50 from scratch combined with the three different classification methods. The bold numbers indicate the highest accuracy in each case }
    \label{tab:resnet_scratch}
    \resizebox{0.69\textwidth}{!}{
    \begin{tabular}{c|ccc}
        \multicolumn{4}{c}{ResNet-50 from scratch}\\
        \hline
        Train Size & \multicolumn{3}{c}{AgrilPlant5} \\
        (\%) & Proposed & OvO\cite{1v1_network2} & OvA\cite{1v1_network2}\\
        \hline
        10 & \textbf{82.27} $\pm$ 1.94 & 77.53 $\pm$ 0.96 & 72.93 $\pm$ 3.85\\
        20 & \textbf{87.07} $\pm$ 2.07 & 85.40 $\pm$ 0.64 & 82.73 $\pm$ 2.29\\
        50 & \textbf{93.93} $\pm$ 1.04 & 91.47 $\pm$ 0.90 & 89.87 $\pm$ 0.77\\
        80 & \textbf{94.73} $\pm$ 0.72 & 93.53 $\pm$ 1.22 & 93.73 $\pm$ 1.50\\
        100 & \textbf{95.60} $\pm$ 0.43 & 94.33 $\pm$ 0.94 & 93.87 $\pm$ 2.06\\
        \hline
        \hline
        Train Size & \multicolumn{3}{c}{AgrilPlant10} \\
        (\%) & Proposed & OvO\cite{1v1_network2} & OvA\cite{1v1_network2}\\
        \hline
        10 & \textbf{76.70} $\pm$ 0.95 & 76.23 $\pm$ 2.06 & 72.93 $\pm$ 2.04\\
        20 & \textbf{87.37} $\pm$ 1.82 & 86.03 $\pm$ 1.29 & 84.20 $\pm$ 1.91\\
        50 & 93.10 $\pm$ 0.45 & 93.13 $\pm$ 0.46 & \textbf{93.20} $\pm$ 0.83\\
        80 & \textbf{96.57} $\pm$ 0.48 & 96.00 $\pm$ 0.53 & 95.03 $\pm$ 1.19\\
        100 & \textbf{97.27} $\pm$ 0.51 & 96.10 $\pm$ 0.38 & 96.23 $\pm$ 0.85\\
        \hline
        \hline
        Train Size & \multicolumn{3}{c}{Tropic5} \\
        (\%) & Proposed & OvO\cite{1v1_network2} & OvA\cite{1v1_network2}\\
        \hline
        10 & \textbf{79.73} $\pm$ 2.72 & 77.31 $\pm$ 1.05 & 73.59 $\pm$ 2.63\\
        20 & 86.98 $\pm$ 1.12 & \textbf{87.41} $\pm$ 3.72 & 83.35 $\pm$ 3.45\\
        50 & \textbf{94.39} $\pm$ 1.44 & 93.47 $\pm$ 2.48 & 91.19 $\pm$ 2.40\\
        80 & 97.24 $\pm$ 0.74 & \textbf{97.29} $\pm$ 1.35 & 96.23 $\pm$ 0.89\\
        100 & 98.26 $\pm$ 0.33 & \textbf{98.64} $\pm$ 0.82 & 97.48 $\pm$ 0.44\\
        \hline
        \hline
        Train Size & \multicolumn{3}{c}{Tropic10} \\
        (\%) & Proposed & OvO\cite{1v1_network2} & OvA\cite{1v1_network2}\\
        \hline
        10 & \textbf{68.80} $\pm$ 2.71 & 67.57 $\pm$ 3.44 & 62.38 $\pm$ 1.42\\
        20 & \textbf{83.46} $\pm$ 1.61 & 82.57 $\pm$ 1.75 & 77.85 $\pm$ 2.10\\
        50 & 93.44 $\pm$ 0.73 & \textbf{93.45} $\pm$ 1.20 & 93.09 $\pm$ 0.76\\
        80 & \textbf{96.90} $\pm$ 0.50 & 96.45 $\pm$ 1.20 & 96.43 $\pm$ 0.88\\
        100 & \textbf{97.68} $\pm$ 0.25 & 97.44 $\pm$ 0.42 & 97.10 $\pm$ 0.57\\
        \hline
        \hline
        Train Size & \multicolumn{3}{c}{Swedish5} \\
        (\%) & Proposed & OvO\cite{1v1_network2} & OvA\cite{1v1_network2}\\
        \hline
        10 & \textbf{90.13} $\pm$ 1.95 & 75.20 $\pm$ 1.96 & 71.76 $\pm$ 1.95\\
        20 & \textbf{94.55} $\pm$ 2.32 & 86.80 $\pm$ 3.26 & 83.53 $\pm$ 1.61\\
        50 & \textbf{99.09} $\pm$ 1.29 & 96.08 $\pm$ 0.95 & 96.48 $\pm$ 1.34\\
        80 & \textbf{99.45} $\pm$ 0.81 & 98.24 $\pm$ 0.83 & 97.92 $\pm$ 0.91\\
        100 & \textbf{99.64} $\pm$ 0.50 & 98.96 $\pm$ 0.46 & 98.72 $\pm$ 0.52\\
        \hline
        \hline
        Train Size & \multicolumn{3}{c}{Swedish10} \\
        (\%) & Proposed & OvO\cite{1v1_network2} & OvA\cite{1v1_network2}\\
        \hline
        10 & \textbf{90.27} $\pm$ 1.39 & 73.52 $\pm$ 3.57 & 63.44 $\pm$ 1.99\\
        20 & \textbf{92.91} $\pm$ 1.59 & 82.32 $\pm$ 4.81 & 83.60 $\pm$ 2.53\\
        50 & \textbf{98.00} $\pm$ 0.76 & 95.56 $\pm$ 0.83 & 95.68 $\pm$ 0.99\\
        80 & \textbf{98.82} $\pm$ 0.69 & 98.00 $\pm$ 0.40 & 97.12 $\pm$ 0.46\\
        100 & \textbf{99.18} $\pm$ 0.59 & 98.40 $\pm$ 0.37 & 98.32 $\pm$ 0.23
    \end{tabular}
    }
\end{table}

\begin{table}[!t]
    \footnotesize
    \centering
    \caption{Classification performance (average accuracy and standard deviation) of the Inception-V3 from scratch combined with the three different classification methods. The bold numbers indicate the highest accuracy in each case }
    \label{tab:inception_scratch}
    \resizebox{0.7\textwidth}{!}{
    \begin{tabular}{c|ccc}
        \multicolumn{4}{c}{Inception-V3 from scratch}\\
        \hline
        Train Size & \multicolumn{3}{c}{AgrilPlant5} \\
        (\%) & Proposed & OvO\cite{1v1_network2} & OvA\cite{1v1_network2}\\
        \hline
        10 & \textbf{82.13} $\pm$ 1.74 & 77.13 $\pm$ 1.28 & 71.67 $\pm$ 2.67\\
        20 & \textbf{88.00} $\pm$ 1.03 & 85.47 $\pm$ 2.10 & 83.33 $\pm$ 3.47\\
        50 & \textbf{94.87} $\pm$ 0.30 & 92.40 $\pm$ 0.86 & 89.73 $\pm$ 1.19\\
        80 & \textbf{96.93} $\pm$ 1.30 & 94.47 $\pm$ 0.90 & 94.33 $\pm$ 0.53\\
        100 & \textbf{97.00} $\pm$ 0.24 & 94.93 $\pm$ 0.37 & 94.80 $\pm$ 1.02\\
        \hline
        \hline
        Train Size & \multicolumn{3}{c}{AgrilPlant10} \\
        (\%) & Proposed & OvO\cite{1v1_network2} & OvA\cite{1v1_network2}\\
        \hline
        10 & \textbf{78.20} $\pm$ 1.20 & 77.80 $\pm$ 3.00 & 73.57 $\pm$ 1.47\\
        20 & \textbf{88.83} $\pm$ 1.20 & 86.97 $\pm$ 1.69 & 85.87 $\pm$ 1.57\\
        50 & \textbf{96.27} $\pm$ 0.38 & 94.87 $\pm$ 1.00 & 94.57 $\pm$ 1.23\\
        80 & \textbf{97.43} $\pm$ 0.51 & 96.47 $\pm$ 0.69 & 96.60 $\pm$ 0.73\\
        100 & \textbf{97.93} $\pm$ 0.38 & 96.90 $\pm$ 0.65 & 97.40 $\pm$ 0.67\\
        \hline
        \hline
        Train Size & \multicolumn{3}{c}{Tropic5} \\
        (\%) & Proposed & OvO\cite{1v1_network2} & OvA\cite{1v1_network2}\\
        \hline
        10 & \textbf{82.48} $\pm$ 2.01 & 82.24 $\pm$ 1.91 & 78.76 $\pm$ 2.09\\
        20 & \textbf{90.61} $\pm$ 0.55 & 89.06 $\pm$ 1.55 & 89.40 $\pm$ 1.47\\
        50 & \textbf{98.11} $\pm$ 0.47 & 97.19 $\pm$ 0.66 & 95.74 $\pm$ 1.15\\
        80 & \textbf{98.89} $\pm$ 0.44 & 98.84 $\pm$ 0.53 & 98.02 $\pm$ 0.47\\
        100 & 99.03 $\pm$ 0.64 & \textbf{99.13} $\pm$ 0.51 & 98.30 $\pm$ 1.06\\
        \hline
        \hline
        Train Size & \multicolumn{3}{c}{Tropic10} \\
        (\%) & Proposed & OvO\cite{1v1_network2} & OvA\cite{1v1_network2}\\
        \hline
        10 & \textbf{77.48} $\pm$ 1.54 & 75.14 $\pm$ 2.73 & 70.46 $\pm$ 3.22\\
        20 & 87.25 $\pm$ 1.01 & 86.77 $\pm$ 1.14 & \textbf{89.43} $\pm$ 2.06\\
        50 & \textbf{96.06} $\pm$ 0.21 & 95.59 $\pm$ 1.28 & 94.78 $\pm$ 0.34\\
        80 & 98.19 $\pm$ 0.30 & \textbf{98.38} $\pm$ 0.70 & 97.42 $\pm$ 0.73\\
        100 & 98.44 $\pm$ 0.40 & \textbf{98.56} $\pm$ 0.46 & 98.54 $\pm$ 0.22\\
        \hline
        \hline
        Train Size & \multicolumn{3}{c}{Swedish5} \\
        (\%) & Proposed & OvO\cite{1v1_network2} & OvA\cite{1v1_network2}\\
        \hline
        10 & \textbf{88.00} $\pm$ 3.82 & 71.60 $\pm$ 4.24 & 66.08 $\pm$ 3.01\\
        20 & \textbf{96.91} $\pm$ 2.09 & 86.40 $\pm$ 2.61 & 86.96 $\pm$ 4.36\\
        50 & \textbf{99.82} $\pm$ 0.41 & 98.40 $\pm$ 0.75 & 95.36 $\pm$ 2.63\\
        80 & \textbf{99.98} $\pm$ 0.04 & 99.36 $\pm$ 0.36 & 98.56 $\pm$ 0.61\\
        100 & \textbf{100.00} $\pm$ 0.00 & 99.76 $\pm$ 0.36 & 99.44 $\pm$ 0.67\\
        \hline
        \hline
        Train Size & \multicolumn{3}{c}{Swedish10} \\
        (\%) & Proposed & OvO\cite{1v1_network2} & OvA\cite{1v1_network2}\\
        \hline
        10 & \textbf{89.82} $\pm$ 1.39 & 79.52 $\pm$ 3.43 & 70.96 $\pm$ 4.19\\
        20 & \textbf{93.02} $\pm$ 1.65 & 91.84 $\pm$ 2.25 & 85.60 $\pm$ 3.90\\
        50 & \textbf{99.09} $\pm$ 0.45 & 97.36 $\pm$ 0.86 & 97.36 $\pm$ 0.96\\
        80 & \textbf{99.91} $\pm$ 0.20 & 99.20 $\pm$ 0.58 & 98.48 $\pm$ 0.39\\
        100 & \textbf{99.91} $\pm$ 0.20 & 99.48 $\pm$ 0.18 & 99.00 $\pm$ 0.51
    \end{tabular}
    }
\end{table}

When the classifiers are trained from scratch, the experiment results favor even more to our proposed model. 

\begin{itemize}
    \item On the AgrilPlant dataset, our proposed model outperforms the other models in almost every setting when using ResNet-50. The advantage is more obvious when incorporated with the Inception-V3 network.
    \item On the Tropic dataset, the two OvO models have fair performance, with our proposed model having a slight advantage. 
    \item On the Swedish dataset, our proposed model has a very significant advantage over the other models. When the training size is low, the difference in accuracy is over $10\%$. When the training size is high, our model achieves over $99\%$ accuracy under all settings.
\end{itemize}

To conclude all the experiment results above, it is observed that our proposed model generally performs better than other models. On the AgrilPlant dataset, the advantage of our proposed model is obvious under most settings. On the Tropic dataset, however, the OvO strategy may be hindered by uneven sample distribution over classes. When trained from scratch, our proposed model generally generates better performance. On the Swedish model, our proposed model has a very significant advantage over other models.

Finally, it is noteworthy that the principal focus of developing the OvO model in \cite{1v1_network2} is to speed up the tedious training process in the conventional OvO strategy, in which each binary classifier is trained from the ground independently. The classification accuracy is not the first issue in their model. In the next experiment, it is demonstrated that by modifying a well-established OvA model into the OvO model, not only the classification accuracy can be improved, but the training process can also be fast following our proposed strategy. 

\subsection{Disease diagnosis}

Multi-class classification plays a crucial role in medical applications, especially disease diagnosis, as it enables healthcare professionals to categorize patients into different disease classes based on their clinical features and symptoms. One application is the electrocardiogram (ECG) analysis.

ECG analysis is a widely used medical technique for diagnosing and monitoring heart conditions by recording and analyzing the heart's electrical signals. It provides crucial information on heart function, including rhythm, rate, and timing. Computer algorithms are used to process and analyze the signals to identify abnormal patterns such as arrhythmias and heart muscle damage.

\begin{figure}[h]
    \centering
    \includegraphics[width=0.6\textwidth]{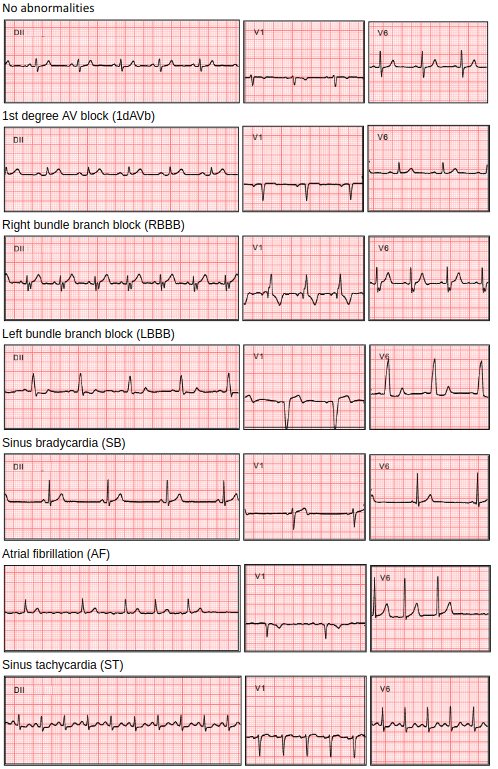}
    \caption{Sample data from each class of the dataset used in \cite{ecg}}
    \label{fig:ecg_data}
\end{figure}

\begin{figure}[h]
    \centering
    \includegraphics[width=\textwidth]{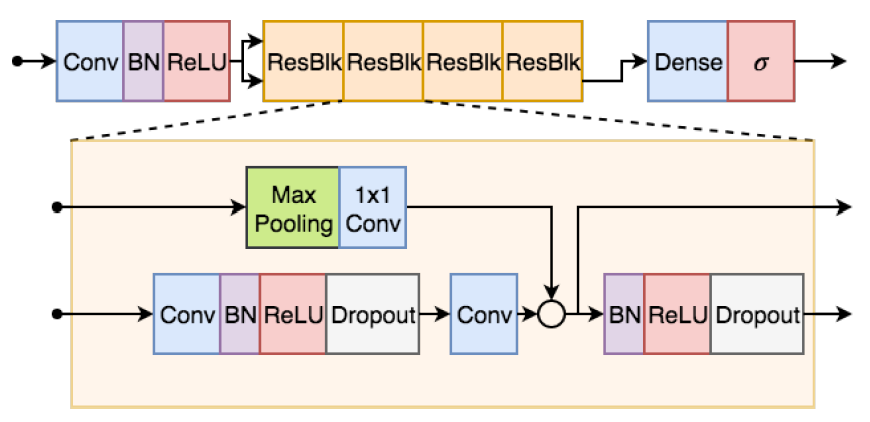}
    \caption{The network architecture of the deep neural network proposed in \cite{ecg}}
    \label{fig:ecg_network}
\end{figure}

In \cite{ecg}, deep neural networks (using OvA strategy) are trained in a dataset consisting of more than 2 million labeled 12-lead ECG data and have been shown to achieve high accuracy in detecting 6 different kinds of abnormalities. Data examples are shown in \Cref{fig:ecg_data} for reference. The network proposed in \cite{ecg} is a modified residual network. The architecture is shown in \Cref{fig:ecg_network} for completeness of the paper. Readers are referred to \cite{ecg} for more details of the network.

In this application, we demonstrate how our method can be used to refine an OvA model into an OvO model by constructing binary classifiers from the existing OvA classification network. From the ECG multi-classification network ${\cal D}$ in \cite{ecg}, our proposed modification to training the binary classifiers ${\cal D}^j$ is applied, where $i,j$ are the classes of different abnormalities. In other words, for each class subset $S_j$ from the dataset $S$, a binary classifier ${\cal D}^j$ is trained from the OvA deep neural network ${\cal D}$. Since the OvA model ${\cal D}$ is already proven to achieve high classification accuracy, it does not take much time to train the classifiers ${\cal D}^j$ before each of them shows high performance.

To train the classifiers, note that a part ($15\%$) of the training data in \cite{ecg} (provided as open-source data) is used to train the classifiers. A joint probability estimate is performed on ${\cal D}^j_i$'s to generate the class label. The testing data presented in \cite{ecg} is also publicly available as open-source data, and is consisted of $827$ samples. The number of positive samples in each class in the training set and that in the testing set are recorded in \Cref{tab:ecg_data_number}.

\begin{table}[h]
    \centering
    \caption{Number of samples for each class in training the binary classifiers and in testing the proposed model}
    \label{tab:ecg_data_number}
    \begin{tabular}{c|cc}
        Class & No. of samples in training & No. of samples in testing \\
        \hline
        1dAVb & 5716 & 28 \\
        RBBB & 9672 & 34\\
        LBBB & 6026 & 30\\
        SB & 5605 & 16\\
        AF & 7584 & 13\\
        ST & 7033 & 37
    \end{tabular}
\end{table}

To evaluate and compare the proposed model with the original network, the classification accuracy of the two models is compared on the above testing dataset, in terms of precision, recall, specificity, and F1 score. The results are recorded in \Cref{tab:ecg_accuracy}. 

\begin{table}[!t]
    \centering
    \caption{Classification performance of the proposed OvO strategy and the OvA strategy as proposed in \cite{ecg} on the 6 classes of the ECG data. The bold numbers indicate the different results of our proposed method compared with the OvA strategy }
    \label{tab:ecg_accuracy}
    \begin{tabular}{c|cccc} 
        \hline
        \multirow{2}{*}{} & \multicolumn{4}{c}{1dAVb} \\
        & Precision (PPV) & Recall (Sensitivity) & Specificity & F1 Score \\
        \hline
        OvA      & 0.867          & 0.929          & 0.995 & 0.897 \\
        Proposed & \textbf{0.871} & \textbf{0.964} & 0.995 & \textbf{0.915} \\
        \hline
        \hline
        \multirow{2}{*}{} & \multicolumn{4}{c}{RBBB} \\
        & Precision (PPV) & Recall (Sensitivity) & Specificity & F1 Score \\
        \hline
        OvA      & 0.895 & 1.000 & 0.995 & 0.944 \\
        Proposed & 0.895 & 1.000 & 0.995 & 0.944 \\
        \hline
        \hline
        \multirow{2}{*}{} & \multicolumn{4}{c}{LBBB} \\
        & Precision (PPV) & Recall (Sensitivity) & Specificity & F1 Score \\
        \hline
        OvA      & 1.000 & 1.000 & 1.000 & 1.000 \\
        Proposed & 1.000 & 1.000 & 1.000 & 1.000 \\
        \hline
        \hline
        \multirow{2}{*}{} & \multicolumn{4}{c}{SB} \\
        & Precision (PPV) & Recall (Sensitivity) & Specificity & F1 Score \\
        \hline
        OvA      & 0.833 & 0.938 & 0.996 & 0.882 \\
        Proposed & 0.833 & 0.938 & 0.996 & 0.882 \\
        \hline
        \hline
        \multirow{2}{*}{} & \multicolumn{4}{c}{AF} \\
        & Precision (PPV) & Recall (Sensitivity) & Specificity & F1 Score \\
        \hline
        OvA & 1.000 & 0.769 & 1.000 & 0.870 \\
        Proposed & 1.000 & \textbf{0.846} & 1.000 & \textbf{0.917} \\
        \hline
        \hline
        \multirow{2}{*}{} & \multicolumn{4}{c}{ST} \\
        & Precision (PPV) & Recall (Sensitivity) & Specificity & F1 Score \\
        \hline
        OvA & 0.947 & 0.973 & 0.997 & 0.960 \\
        Proposed & \textbf{0.973} & 0.973 & \textbf{0.999} & \textbf{0.973} \\
        \hline
    \end{tabular}
\end{table}

From the results, our proposed OvO strategy is shown to help improve the classification accuracy of the original OvA model on 3 classes, i.e. 1dAVb, AF and ST. The two models show no difference in terms of accuracy on the other 3 classes, i.e. RBBB, LBBB and SB. This can be explained by the implementation of the OvO strategy which withdraws most normal control data in the training process to increase the discriminating power of the classifiers. Moreover, the joint probability estimate combines the pairwise accuracy to provide an accurate class label probability. Altogether the strategy helps to improve the well-established OvA classification model to generate even more accurate results.

In medical applications, data imbalance is almost inevitable as there are limitations in collecting patients' data. Leveraging with an existing and well-established OvA classification model, the proposed OvO modification is helpful to balance the discriminating power of each binary classifier and therefore gives a more accurate classification. Moreover, such modification does not require training a lot of classifiers from scratch, and hence diminishes the time cost as much as possible. 

\section{Conclusion}
In this paper, a One-versus-One (OvO) deep learning model incorporating joint probability measure and calibration is proposed for solving the multi-classification problem. The proposed model employs a distance measurement on the separating feature hyperplane to calibrate the pairwise probability of the binary classifiers. The calibrated pairwise probability measures are combined by a joint probability measure to obtain the class probability, which is implemented by solving a linear system related to the Kullback-Leibler distance minimization problem. 
% The major merit of the proposed model is on the implementation of the joint probability measure being more sophisticated than the conventional voting mechanism to combine the binary classifiers. 
The major merit of the proposed model is the use of more sophisticated joint probability measures for better classification accuracy, rather than the conventional voting mechanism to combine the binary classifiers. The proposed model can be applied in both cases when the binary classifiers are to be built from scratch, or to be built by modifying existing multi-classification networks.
The proposed model is compared with state-of-the-art models on multiple datasets and is shown to achieve higher classification accuracy under most settings. It is evident that the proposed model is beneficial in many real applications.

\section*{Acknowledgment}

The work of R. Chan was supported in part by HKRGC Grants Nos. CUHK14301718, NCityU214/19, CityU11301120, CityU11309922, C1013-21GF, and CityU Grant9380101.

R. Chan is with the Department of Mathematics, City University of Hong Kong, Tat Chee Avenue, Kowloon Tong, Hong Kong SAR, China; Hong Kong Centre for Cerebro-Cardiovascular Health Engineering, 
Email: raymond.chan@cityu.edu.hk 

A. Chan is with the Hong Kong Centre for Cerebro-Cardiovascular Health Engineering, Email: hlchan@hkcoche.org

L. Dai is with the Department of Mathematics, City University of Hong Kong, Tat Chee Avenue, Kowloon Tong, Hong Kong SAR, China,
Email: ljdai2-c@my.cityu.edu.hk

\bibliographystyle{IEEEtran}
\bibliography{ref}

\end{document}